\documentclass[10pt,twocolumn,letterpaper]{article}

\usepackage{times}
\usepackage{epsfig}
\usepackage{graphicx}
\usepackage{amsmath}
\usepackage{amssymb}
\usepackage{subfig}
\usepackage{tabularx}
\usepackage{tikz}
\usetikzlibrary{scopes,arrows,decorations.pathmorphing,backgrounds,positioning,fit,petri, trees}

\usepackage[pagebackref=true,breaklinks=true,letterpaper=true,colorlinks,bookmarks=false]{hyperref}

\begin{document}

\title{A Continuous, Full-scope, Spatio-temporal Tracking Metric based on KL-divergence}

\author{Terry Adams \\
U.S. Government\\
{\tt\small terry@ieee.org}
}

\maketitle

\begin{abstract}
A unified metric is given for the evaluation of object tracking systems.  
The metric is inspired by KL-divergence 
or relative entropy, which is commonly used to evaluate clustering techniques.  Since tracking problems 
are fundamentally different from clustering, the components of KL-divergence are 
recast to handle various types of tracking errors (i.e., false alarms, missed detections, merges, splits). 
Scoring results are given on a standard 
tracking dataset (Oxford Town Centre Dataset), as well as several simulated scenarios. 
Also, this new metric is compared with several other metrics including the commonly used 
Multiple Object Tracking Accuracy metric. 
In the final section, advantages of this metric are given including the fact 
that it is continuous, parameter-less and comprehensive. 
\end{abstract}


\section{Introduction}
In 1951, S. Kullback and R.A. Leibler introduced a measure of divergence or discrimination between two distributions \cite{KL1951}. 
It has since been referred to as Kullback-Leibler (KL) divergence, and  been used in many areas of statistics, machine learning and applied neuroscience. 
Its formulation is based on conditional entropy and Shannon's entropy.  It enjoys nice properties such as being additive for 
independent distributions. 

By itself, KL-divergence is not a metric, but the symmetrized version is a metric and induces a topology on the space of probability distributions. 
It has commonly been used as a metric for clustering algorithms.  In this setting, it is often referred to as relative entropy. 
Given a single reference clustering $Q$, then the error of a computed clustering $P$ can be measured as
\[
H(P|Q) + H(Q|P) . 
\]
This error will evaluate to zero, if $P = Q$, and equals $H(P) + H(Q)$, if $P$ is independent of $Q$. 
%

For many problems in computer vision, or more broadly in statistical estimation, 
the goal is to track or detect the presence 
of an object or event. 
See the references 
\cite{LDZZMHJMS2018, GBSHS2014, BES2006, BS2008, MTRRS2016, PHSB2006, Sen2001, Ell2002, Brown2005, BP2006, YMV2007, ZSTWWT}, 
as well as the citations contained therein for a background on tracking metrics. 
In the paper \cite{KDH2009}, the authors use relative entropy and mutual information 
in a different manner to produce a tracking metric. 
In the following subsection, we give details on the major 
differences between our metric here, and the metric described in \cite{KDH2009}. 

The tracking problem is fundamentally different from the clustering problem, 
in that, it can be an open universe problem.  
Often in clustering, there is a well-defined superset of objects 
or elements that are being clustered.  From this closed set, a probability space may be established. 
This makes the problem of clustering tractable for probabilistic methods. 
However, the problem of tracking can take place in a very large universe of possible tracks. 
In particular, if the goal is to track a person in space and time from one or more video sequences, 
the set of possible tracks becomes vast.  For high-def video at 30 frames per second, 
a spatio-temporal track may include billions of pixels. 
Still, we are able to extend the probabilistic 
notion of divergence to a unified metric for evaluating tracking problems, 
as well as event detection. 
This has the benefit that notions such as track fragmentation and track purity can be handled 
with a single metric, along with false alarms or partial missed detections.  See \cite{PHSB2006} 
for a definition of track fragmentation and track purity. 

In order to extend the notion of divergence to the tracking regime, we separate the standard 
formula for the symmetrized KL-divergence (and cross entropy) into multiple parts. 
The component parts from the KL formulas naturally map to various types of errors. 
In all, we identify three main types of errors, referred to as:
\begin{enumerate}
\itemsep-.10em 
\item Inner divergence,
\item Outer divergence,
\item Track density divergence.
\end{enumerate}
The inner divergence error represents error obtained from splitting tracks 
or merging tracks. 
The outer divergence error measures error produced from false alarms or missed detections. 
Also, included is a track density divergence which measures error introduced 
from duplicate system outputs, and more generally the difference between densities 
produced from system tracks and densities produced from ground-truth tracks. 
In a statistical sense, these three classes of error can be rolled 
up into a single metric for measuring tracking performance. 

Prior to defining the complete metric, we give further heuristics for this KL-inspired 
tracking metric that associate common tracking error types with the corresponding 
components of the KL-divergence formulas.  Suppose that 
$\mathcal{T} = \{\tau_1, \ldots , \tau_n\}$ is a set of ground-truth 
tracks, and $\mathcal{S}  = \{\sigma_1, \ldots , \sigma_m \}$ is a set of system tracks. 
At the moment, assume that there is no intersection between ground-truth tracks, 
and likewise, there is no intersection between system tracks. 
Place a probability measure $\nu$ on a given set of tracks in the following manner. 
Assume $\nu$ is defined on $\mathcal{T}$ where each track is given equal weight, ${1} / {n}$. 
Each track $\tau_i$ may be thought of as a spatio-temporal volume. 
Define $\nu$ to give uniform mass on $\tau_i$ based on the number of pixels in $\tau_i$. 
For this paper, a pixel is indexed by both the spatial location and the frame or timing information. 
Thus, two pixels are identical (in position), if the spatial location is the same and 
the frame index is identical.  Otherwise, two pixels are disjoint.  For this paper, 
pixels can be thought of as atomic elements which compose entire tracks. 
The formula for symmetrized KL-divergence, or more directly, for cross entropy, is 
\[
H( \mathcal{S} | \mathcal{T} ) + H( \mathcal{T} | \mathcal{S} )  
\]
where $H( * | * )$ represents standard conditional entropy using 
the measure $\nu$. 
In our setting, this total is referred to as the total inner divergence. 
Splitting errors are associated with $H( \mathcal{S} | \mathcal{T} )$, 
and merging errors are associated with $H( \mathcal{T} | \mathcal{S} )$. 

The outer divergence is defined in a similar framework as the inner divergence, 
but accounts for false alarms and missed detections. 
We define another error component we call track density divergence. 
This error penalizes spurious duplicate system outputs, or in general, 
how the system tracks differ in distribution conidtionally 
from the ground-truthed tracks (and vice-versa). 

\section{Related Work}
Tracking is a long studied field, whether focused specifically 
on object tracking in video or tracking from other signals, 
there is a large body of research in this area. 
There is no single survey that gives an exhaustive 
account of all prior research. Our approach here 
is to start by listing the most popular challenges 
on video tracking or re-identification in video. 
Each one of these challenges has a primary metric. 
Across all video tracking challenges, a single metric 
has emerged as the most used metric. 
It is called Multiple Object Tracking Accuracy (MOTA). 
It's defined in section \ref{mota-define} 
and later in section \ref{mota-compare}, 
we give direct comparisons 
between our KL-based metric and MOTA, as well as 
other established metrics. 

We also describe some other lesser known 
metrics that might appear similar in spirit 
to the new KL-based metric introduced here. 
However, we are able to show several major differences between our 
KL-based metric and these previously defined information 
theoretic metrics. 

\subsection{Tracking challenges}
There are several challenges that address person or object 
tracking from video. Also, recently, there are several 
new datasets that are used for tracking and re-identification. 
Table \ref{table:CTC-1} lists the challenge or dataset along with 
the primary metric used for that challenge set. 
CMC stands for Cumulative Matching Characteristic 
and mAP stands for mean Average Precision. 
\begin{table}
\footnotesize 
\begin{tabular}{ lc }
Challenge & Primary Metric\\
\hline 
Multiple Object Tracking challenge & MOTA \\
Performance Evaluation of Tracking and Surveillance & MOTA \\
Duke Multi-Target, Multi-Camera Tracking Project & MOTA \\
TRECVID Media Event Detection & mAP \\
TRECVID Surveillance Event Detection & MOTA \\
Chinese University of Hong Kong Re-id & CMC \\
Market-1501 Re-id & mAP \\
Multi-Scene Multi-Time Re-id & CMC \\
\end{tabular}
\caption{Common Tracking Challenges}
\label{table:CTC-1}
\end{table}

\subsection{MOTA}
\label{mota-define}
The Multiple Object Tracking Accuracy (MOTA) metric is the most commonly 
used metric for well established tracking challenges. 
MOTA is defined on a per frame basis. 
For each frame $t$, let 
$m_t$ be the number of misses, 
$fp_t$ be the number of false positives, 
$\varPhi_t$ represents the number of ID switches, 
and $g_t$ be the number of true detections for frame $t$.
Define 
\[
MOTA = 1 - \frac{ \sum_{t} \big( m_t + fp_t + \varPhi_t \big) }{ \sum_t g_t} 
\]
\vskip .2in
The Hungarian algorithm \cite{Kuhn1955} is used to match ground-truth boxes 
to system boxes for each frame. 
A system track stays assigned to the same ground-truth track, 
unless  the IoU
\footnote{ IoU stands for Intersection over Union. Often, detectors or trackers apply
a threshold for correct detections such as $\{\mbox{ IoU} \geq 0.5 \}$. } 
drops below 0.5 for the current frame. 

\subsection{Comparison to previously defined information-theoretic metrics}
There have been several previously defined information-theoretic metrics 
including \cite{MGR1973, SM1992, Raz1992, LNP2004, SVV2008, KDH2009}. 
However, these metrics do not handle all of the scenarios involved 
in spatio-temporal tracking and do not include the list of properties shown 
in Table \ref{adv}. 

At first glance, the tracking metric defined in \cite{KDH2009} may appear 
similar to the metric defined here.  However, there are major differences. 
The metric described in \cite{KDH2009} relies on an association between truthed tracks 
and system tracks, and includes preset thresholds. 
The metric defined in this paper does not require defining an association 
between system tracks and truthed tracks.  Also, our metric does not 
require pre-defined thresholds to calculate errors. 
Thus, the overall error measure is continuous with respect 
to the system tracks. 
Another major difference 
with the tracking metric described here is that our metric 
incorporates both spatial and temporal overlap to produce 
an error based on track volumes. 
The metric in \cite{KDH2009} 
does not include a spatial element, and measures overlap temporally. 
Also, the interpretation 
of $H(S|T)$ as information in the system tracks ($S$) that is not in the 
truth tracks ($T$), does not necessarily capture error produced from false alarms. 
This is influenced heavily by the joint distribution imposed on the 
$\emptyset$/$\emptyset$ outcome for $S$ and $T$. 
This is the space outside $S$ tracks and $T$ tracks. 
The authors mention a heuristic way for measuring the unknown distribution 
outside the truth and system tracks.  It is largely based on the 
size of the sample space of possible tracks. 
However, this might be estimated to be very large and wash out the 
conditional probabilities involving $S$ and $T$ tracks. 
One of our insights is that we can use the range of errors for the inner 
divergence to help scale the error formula for the outer divergence 
(i.e., false alarms, missed detections). 

There are several metrics based on measuring the difference 
between two distributions, or more specifically, the distance 
between two different sets of points. 
Examples include the Hausdorff distance or mass transfer metrics 
such as the Wasserstein metric. However, these metrics have 
drawbacks in the case of video tracking. The paper 
\cite{SVV2008} highlights several of the drawbacks. 
Also, the references \cite{SVV2008} and \cite{RVCV2011} 
propose an information theoretic metric, 
called Optimal SubPattern Assignment (OSPA), 
that improves on the previously defined mass transfer metrics. 
Nevertheless, OSPA does not satisfy three of our main requirements 
for a video tracking metric including:
\begin{itemize}
\item OSPA relies on a one-to-one assignment given by a permutation.
\item OSPA does not handle split or merged tracks gracefully.
\item OSPA depends a threshold (referred to as $c$).
\end{itemize}
Note, KL-divergence is also used to measure the difference 
between different probability distributions. 
However, we place effort in adapting KL-divergence or relative 
entropy to the problem of tracking through a detailed 
look at the component errors that surface when evaluating 
trackers. These component errors are splitting errors, merging errors, 
false alarms, missed detections and spurious duplicate detections 
or misses. Now let's proceed with precisely defining the metric.

\section{Components of Track Divergence}
In this section, we define the individual components that make up the total divergence 
between two spatio-temporal tracks. 
Given a video frame, a subframe $S$ is a collection of pixels from the frame.  
This collection can be thought of as a 2-dimensional region with area $a(S)$. 
A video track $\tau$ is a sequence of subframes $S_1, S_2, \ldots , S_k$ (also known as a tuboid).  
It has a volume $v(\tau)$ defined as 
$v(\tau) = \sum_{i=1}^{k} a(S_i)$. 
To view the formulas in a probilistic framework, the measure is normalized 
to assign equal weight to each track.  In particular, if $\tau$ is one of $n$ 
ground-truth tracks, then 
\[
v(\tau) = \sum_{i=1}^{k} a(S_i) = \frac{1}{n} . 
\]

\subsection{Inner divergence}
The logarithm function (log) is used throughout this paper. 
In many places, the base of the logarithm is not important, 
although, it may be assumed that $\log$ is computed using base 2 
(i.e., $\log{(2)} = 1$). 
The inner divergence of track $\tau$ from track $\sigma$ is computed 
in the following manner:
\[
D_{id} ( \tau | \sigma ) = - \frac{v(\tau \cap \sigma)}{v(\sigma)} 
\log{\frac{v(\tau \cap \sigma)}{v(\sigma)}} 
\]
Given a set of tracks $\mathcal{T}$ and single track $\sigma$, 
\begin{eqnarray*}
D_{id} (\mathcal{T} | \sigma) = \sum_{i=1}^{n} D_{id} (\tau_i | \sigma) . 
\end{eqnarray*}
Provided two sets $\mathcal{T} = \{\tau_1, \ldots , \tau_n\}$ and 
$\mathcal{S}  = \{\sigma_1, \ldots , \sigma_m \}$ of tracks, 
the inner divergence of $\mathcal{T}$ from $\mathcal{S}$ is 
\begin{eqnarray*}
D_{id} (\mathcal{T} || \mathcal{S}) = \frac{1}{m} \sum_{j=1}^{m} D_{id} ( \mathcal{T} | \sigma_j ) . 
\end{eqnarray*}

\subsection{Outer divergence}
Given a set of tracks $\mathcal{T} = \{ \tau_1, \tau_2, \ldots, \tau_n \}$ 
and single track $\sigma$, let 
\[
\alpha =  \frac{ v \big( \sigma \cap (\bigcup_{i=1}^{n} \tau_i) \big) }{ v(\sigma) } . 
\]
Define the outer divergence of $\mathcal{T}$ from track $\sigma$ as: 
\begin{eqnarray*}
D_{od} ( \mathcal{T} | \sigma ) = \log{ \Big( \frac{2 + n}{1 + \alpha (1 + n)} \Big) } . 
\end{eqnarray*}
The outer divergence of track set $\mathcal{T}$ from track set $\mathcal{S}$ 
is defined as,
\begin{eqnarray*}
D_{od} (\mathcal{T} || \mathcal{S}) = \frac{1}{1+m} \sum_{j=1}^{m} D_{od} (\mathcal{T} | \sigma_j) . 
\end{eqnarray*}

\subsubsection{Outer Divergence Heuristics}
Intuitively, the outer divergence formula is based on the worst case scenario for the inner divergence. 
In particular, if a single system track $\sigma$ is covered disjointly by $n$ reference tracks 
$\tau_i$ for $1\leq i\leq n$, then the inner divergence is maximized if each reference 
track intersects $\sigma$ by exactly ${1} / {n}$ of $\sigma$. 
I.e., $\nu (\tau_i \cap \sigma) = ({1} / {n}) \nu(\sigma)$. 
If we were to distribute the mass from $n$ reference tracks to $n$ system tracks, then 
inner divergence is maximized again with a uniform distribution. 
This is based on the fact that the entropy formula for $H(P)$ is maximized, when a partition 
$P$ has uniform weights ${1} / {n}$. 
This maximum value for a partition into $n$ pieces is 
$- \sum_{i=1}^{n} {1} / {n} \log{({1} / {n})} = \log{n}$. 
When there is no intersection ($\alpha = 0$), the outer divergence formula 
produces a value on the same order of magnitude at $\log{(2+n)}$. 
When $\alpha = 1$, the formula produces:
\begin{eqnarray*}
D_{od} ( \mathcal{T} | \sigma ) &=& \log{ \Big( \frac{2 + n}{1 + \alpha (1 + n)} \Big) } \\ 
&=& \log{ \Big( \frac{2 + n}{2 + n} \Big) } = \log{1} = 0 . 
\end{eqnarray*}
The use of the fraction inside the $\log$ formula helps to keep the error bounded 
as the number of reference tracks grow.  
(See section \ref{bes-subsection} for a specific case demonstrating this property.) 
This helps to make error measurements 
comparable across different evaluation datasets. 
Also, note the use of ${1} / {(1+m)}$ in the formula 
for $D_{od}(\mathcal{T}||\mathcal{S})$ makes 
$D_{od}(\mathcal{T}||\mathcal{S})$ well defined when $m=0$. 
Similarly, the metric is well defined if there are no ground truth tracks ($n=0$). 

\subsubsection{Experiments with various averaging methods}
\label{exp-with-means}
We experimented with several different formulas for computing the mean 
in the outer divergence formula. In general, the outer divergence formula 
may be written as, 
\begin{eqnarray*}
D_{od} (\mathcal{T} || \mathcal{S}) = \frac{1}{\mu(m,n)} \sum_{j=1}^{m} D_{od} (\mathcal{T} | \sigma_j) , 
\end{eqnarray*}
where $\mu = \mu(m,n)$ is computed as a type of mean between $m$ and $n$. 
For these experiments, we looked at the VIRAT data \cite{VIRAT-url} 
and considered a complex sequence of 17,070 frames with multiple movers. 
The ground truth contains 43 tracks.  We compared this to a single system output 
of 685 tracks.  Also, we added two (2) tracks randomly to the ground truth, as well 
as created a third pseudo ground truth file with four (4) more tracks that do no overlap 
any of the system tracks.  Typical tracking metrics will assign a high error, 
since there is a large discrepancy between the number of system tracks and ground 
truth tracks.  Errors will be dominated by a large number of false alarms. 
Since the union of the system tracks match the ground truth tracks relatively well, 
although the ground truth tracks are split into much smaller pieces, 
the errors produced by our KL-divergence formulas do not recklessly diverge to infinity. 

\begin{table}
\footnotesize 
\begin{tabular}{ lccc }
Avg Type ($\mu$) & Original GT & Pseudo GT-45 &  Pseudo GT-49\\
\hline 
1 + $n$ & 3.642006 & 3.564423 & 3.327801\\
{\bf 1 + m} & {\bf 3.225075} & {\bf 3.526724} & {\bf 3.653328}\\
arithmetic & 2.775539 & 2.751111 & 2.63099\\
harmonic & 3.43354 & 3.545573 & 3.490565\\
geometric & 2.987377 & 3.010258 & 2.918232\\
\end{tabular}
\caption{VIRAT Experiments Across Different Means}
\label{table:ODMeans}
\end{table}

Using the mean $\mu(m,n) = 1 + m$ produced the most consistent results. 
See Table \ref{table:ODMeans} for the results from running these experiments. 
This makes sense, since the summand contains $m$ elements. 
The github software for ganitametrics reflects this formula. 

\subsection{Track density divergence}
If a system outputs the same ground-truth track twice, while the track occurs only once 
in the ground-truth, then we expect an error to be generated.  The distance between these sets 
of tracks should not be zero. 
Observe that neither the inner divergence nor the outer divergence, as described previously, penalize 
a system for outputting the same ground-truth track multiple times.  The track density divergence 
described here assigns a penalty or error for generating spurious duplicative tracks.

Kullback-Leibler divergence is a common method for measuring the difference between two distributions. 
When comparing two discrete distributions, $P = \{ p_1, p_2, \ldots ,p_n\}$ and 
$Q = \{ q_1, q_2, \ldots , q_n\}$, 
\[
D_{KL}(P || Q) = \sum_{i=1}^{n} p_i \log{ \big( \frac{p_i}{q_i} \big) } . 
\]
For this formula, it is assumed that $P$ is absolutely continuous with respect to $Q$. 
I.e., $p_i = 0$ whenever $q_i = 0$.  This assumption fits well with our needs, since 
the outer divergence measures errors when the two distribuitons do not overlap. 
Here we will compute the KL-divergence relative to each track, and then average 
this across all tracks.  To make a symmetrized version, this is done relative 
to ground-truth tracks, and system tracks separately. 

For each pixel location $x$, define 
\[
\mathcal{T}(x) = \sum_{i=1}^{n} I_{\tau_i}(x) 
\]
where $ I_{\tau_i}(x)$ is the indicator function for $\tau_i$. 
For $x \in \tau_i$, let 
\begin{eqnarray*}
\mathcal{S}_{i}(x) = \sum_{j=1}^{m} I_{\sigma_j}(x) I_{\tau_i}(x) . 
\end{eqnarray*}
Define 
\[
N( \tau_i ) = \sum_{x \in \tau_i} \mathcal{S}_{i}(x) 
\]
and 
\[
\tau_i^{*} = \{ x \in \tau_i : \mathcal{S}_i(x) > \mathcal{T}(x) \} . 
\]
Let the track density divergence of $\mathcal{S}$ given track $\tau_i$ be 
\begin{eqnarray*}
D_{td}( \mathcal{S} | \tau_i ) = \frac{1}{N(\tau_{i})} \sum_{x \in \tau_i^{*}} 
S_{i}(x) \log{ \big( \frac{\mathcal{S}_{i}(x)}{\mathcal{T}(x)} \big) } 
\end{eqnarray*}
and the total track density divergence given the track set $\mathcal{T}$:
\begin{eqnarray*}
D_{td}( \mathcal{S} || \mathcal{T} ) &=& \frac{1}{n} \sum_{i=1}^{n} D_{td} (S | \tau_i) \\
&=& \frac{1}{n} \sum_{i=1}^{n} 
\bigg( \frac{1}{N(\tau_{i})} \sum_{x \in \tau_i^{*}} 
S_{i}(x) \log{ \big( \frac{\mathcal{S}_{i}(x)}{\mathcal{T}(x)} \big) } \bigg)  .
\end{eqnarray*}

\section{Total Track Divergence}
The total track divergence is obtained as a sum of the inner divergence, outer divergence and 
track density divergence.  We include the symmetrized version for each of these; 
thus, the error measurement becomes the addition of six terms:
\begin{eqnarray*}
D_{TD} (\mathcal{T}, \mathcal{S}) &=& D_{id}( \mathcal{T} || \mathcal{S} ) + D_{id}( \mathcal{S} || \mathcal{T} ) \\
&+& 
D_{od}( \mathcal{T} || \mathcal{S} ) + D_{od}( \mathcal{S} || \mathcal{T} ) \\
&+& 
D_{td}( \mathcal{T} || \mathcal{S} ) + D_{td}( \mathcal{S} || \mathcal{T} ) 
\end{eqnarray*}
This is used only when individual tracks within $\mathcal{T}$ are disjoint, and individual tracks 
within $\mathcal{S}$ are disjoint. 
In the following section, it is shown how to handle track sets containing overlapping tracks. 

\subsection{Purified track divergence}
Notice that if the reference tracks $\mathcal{T}$ have overlaps within individual tracks, 
then  $D_{TD}(\mathcal{T}, \mathcal{T}) > 0$.  
In particular, $D_{id}(\mathcal{T}, \mathcal{T}) > 0$.
To produce a divergence that assigns 0 error, when a system outputs reference tracks, 
define purified inner divergence as, 
\begin{eqnarray*}
D_{pid}(\mathcal{T} || \mathcal{S}) = \left\{
\begin{array}{l}
                  0, \ \mbox{ if $D_{id}(\mathcal{T} || \mathcal{T}) > D_{id}(\mathcal{T} || \mathcal{S})$ },  \\
                  D_{id}(\mathcal{T} || \mathcal{S}) - D_{id}(\mathcal{T} || \mathcal{T}), \ \mbox{otherwise} . 
                \end{array}
              \right.
\end{eqnarray*}
Thus, for any track set $\mathcal{T}$, 
\[
D_{pid} (\mathcal{T} || \mathcal{T}) = 0 . 
\]

\subsection{General formula for the KL-inspired tracking metric}
\label{gen-for-subsection}
Given any two sets of tracks $\mathcal{T}$ and $\mathcal{S}$, define 
\begin{eqnarray*}
D_{TD} (\mathcal{T}, \mathcal{S}) &=& D_{pid}( \mathcal{T} || \mathcal{S} ) + D_{pid}( \mathcal{S} || \mathcal{T} ) \\
&+& 
D_{od}( \mathcal{T} || \mathcal{S} ) + D_{od}( \mathcal{S} || \mathcal{T} ) \\
&+& 
D_{td}( \mathcal{T} || \mathcal{S} ) + D_{td}( \mathcal{S} || \mathcal{T} ) 
\end{eqnarray*}

\section{Results on the Oxford Town Centre Tracking Data}
The Active Vision Laboratory at Oxford University has collected a rich video dataset 
for evaluating pedestrian tracking algorithms.  
It is publicly available at \cite{BR-url}. 
The dataset contains reference tracks which give highly accurate bounding boxes for all movers 
(people, strollers).  Also, provided are outputs from two systems, one described in a 2009 BMVC paper, 
and another described in a 2011 CVPR paper. 

To demonstrate our metrics, we implemented them in a C++ library called GanitaMetrics. 
It includes a driver program (gmetrics) for calling a few different functions. 
Table \ref{table1} and Table \ref{table2} display 
output from running on the two different system outputs.  
The total error is a sum of six component errors as presented in section \ref{gen-for-subsection}. 
Tables \ref{table1} and \ref{table2} show the labels for each of these six component errors. 
Note, the BMVC results 
generate a better score than the CVPR 2011 output. 
\begin{table}
\centering
\small 
\begin{tabular}{lcc}
Reference \# of Tracks & 230 \\
System    \# of Tracks & 455 \\
Inner div relative to reference 	& 0.758774  & $D_{pid}(\mathcal{S} || \mathcal{T})$ \\
Inner div relative to system    	& 0.191656  & $D_{pid}(\mathcal{T} || \mathcal{S})$ \\
Total inner div error           	& +0.950429 \\
Missed detection error          	& +0.508036 & $D_{od}(\mathcal{S} || \mathcal{T})$ \\
Missed detection proportion     	& 0.312293  \\
Density error rel to reference  	& +0.025399 & $D_{td}(\mathcal{S} || \mathcal{T})$ \\
False alarm error               	& +1.304291 & $D_{od}(\mathcal{T} || \mathcal{S})$ \\
False alarm proportion          	& 0.216210  \\
Density error rel to system     	& +0.023201 & $D_{td}(\mathcal{T} || \mathcal{S})$ \\
                                        &-----------\\
Total KL-track error            	& =2.811357 \\
\end{tabular}
\caption{Tracking Error Summary for CVPR 2011}
\label{table1}
\end{table}

\begin{table}
\centering
\small
\begin{tabular}{lcc}
Reference \# of Tracks & 230 \\
System    \# of Tracks & 244 \\
Inner div relative to reference 	& 0.285220 & $D_{pid}(\mathcal{S} || \mathcal{T})$  \\
Inner div relative to system    	& 0.444053  & $D_{pid}(\mathcal{T} || \mathcal{S})$  \\
Total inner div error           	& +0.729273 \\
Missed detection error          	& +1.275770 & $D_{od}(\mathcal{S} || \mathcal{T})$ \\
Missed detection proportion     	& 0.426925  \\
Density error rel to reference  	& +0.008601 & $D_{td}(\mathcal{S} || \mathcal{T})$ \\
False alarm error               	& +0.327670 & $D_{od}(\mathcal{T} || \mathcal{S})$ \\
False alarm proportion          	& 0.176251  \\
Density error rel to system     	& +0.033870 & $D_{td}(\mathcal{T} || \mathcal{S})$ \\
                                        &-----------\\
Total KL-track error            	& =2.375185 \\
\end{tabular}
\caption{Tracking Error Summary for BMVC 2009}
\label{table2}
\end{table}
The CVPR 2011 system outputted 455 tracks whereas ground-truth contains 230 tracks.  
This leads to a higher false alarm error as noticed by 1.304291, 
and a higher "inner div relative to reference": 
0.758774 compared to 0.285220.

\begin{figure}
\centering
\subfloat[Ground-truth annotations \cite{BR-url}]{\includegraphics[width=7cm]{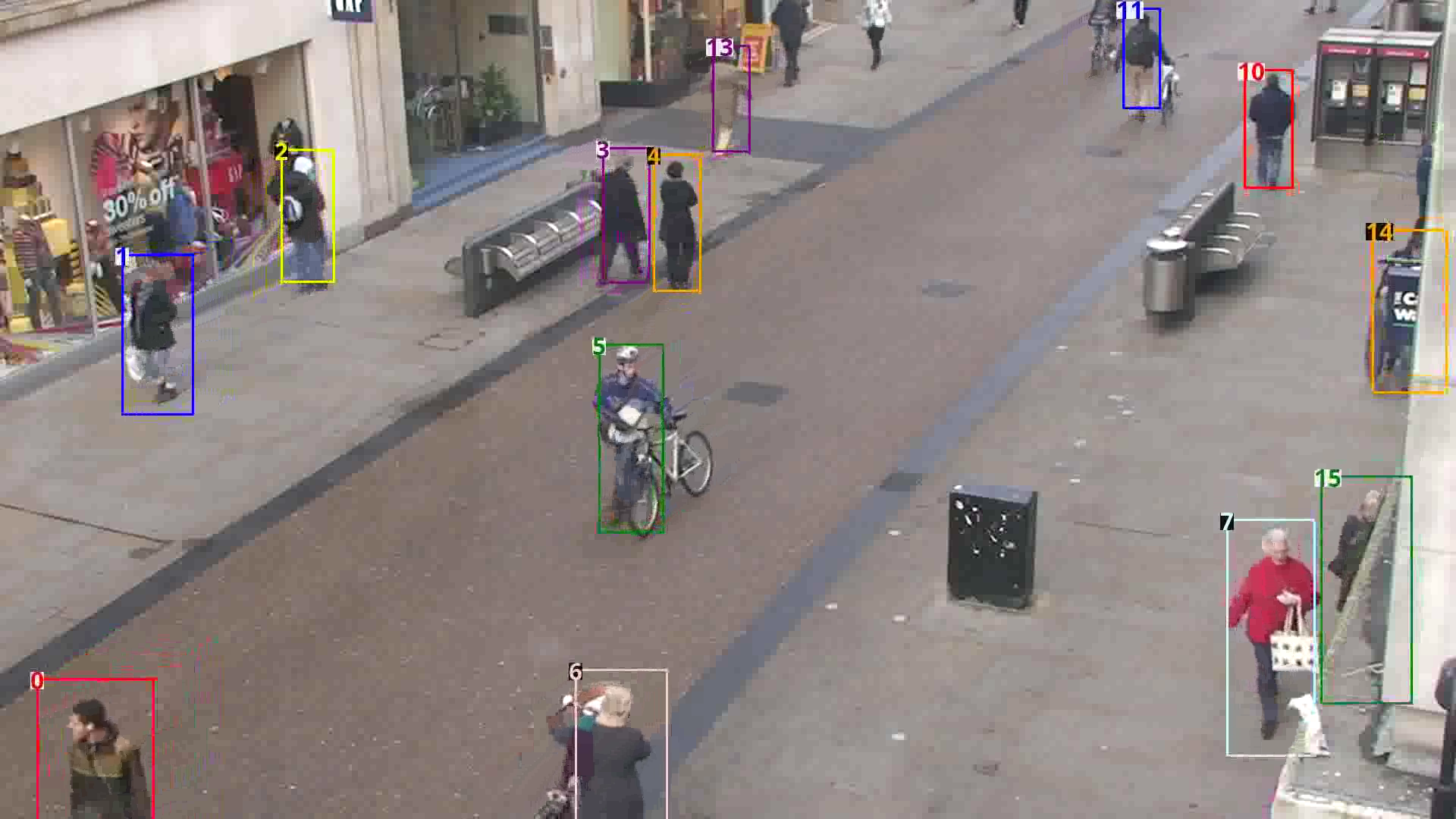}}%
\qquad
\subfloat[Boxes output by BMVC 2009 system]{\includegraphics[width=7cm]{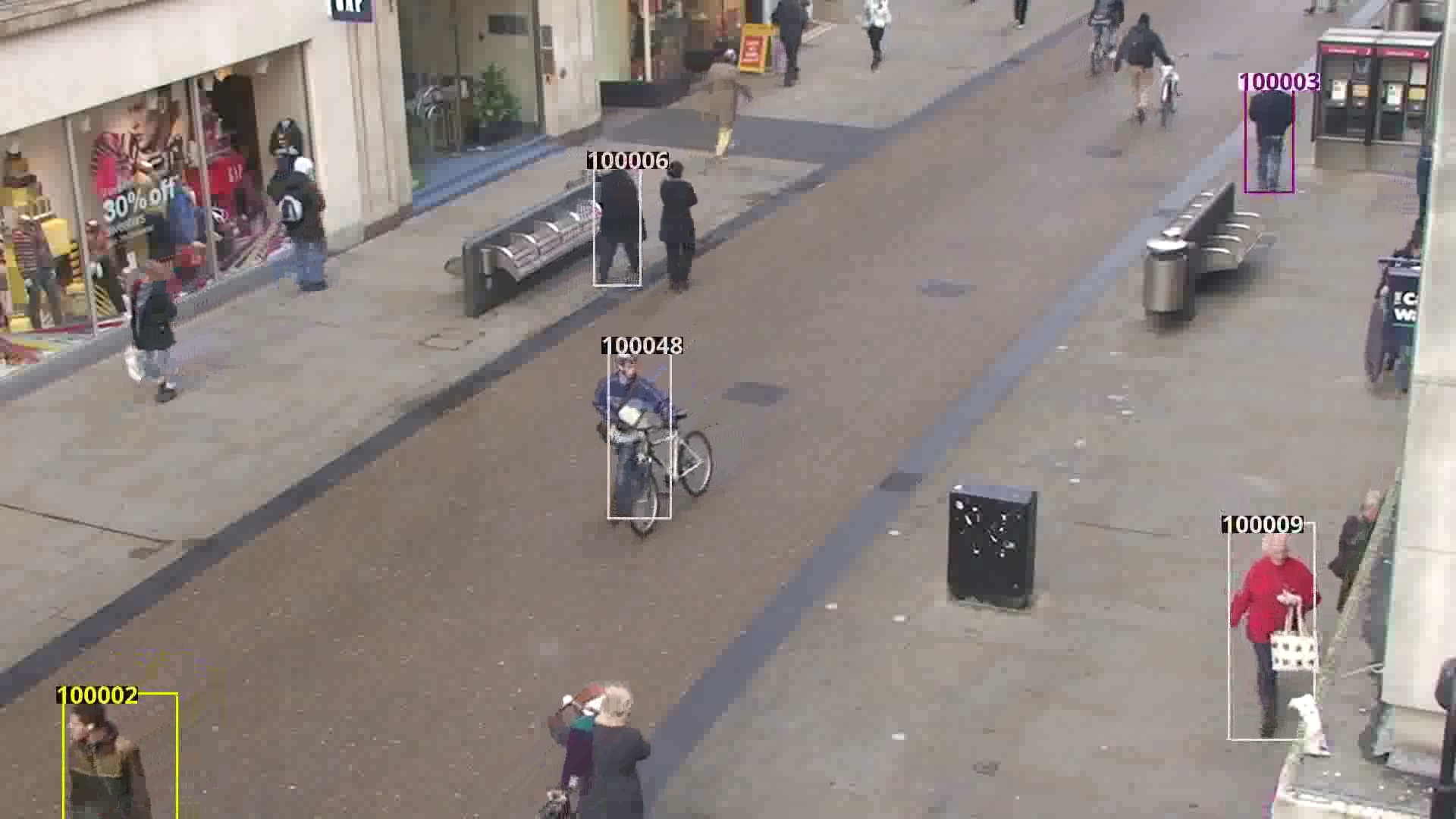}}%
\caption{Visualization of boxes from gmetrics}
\label{fig:TC-GT}
\end{figure}

The gmetrics program includes a visualization option that outputs 
a script that can generate a video with overlaid boxes. 
See Figure \ref{fig:TC-GT} for a comparison of ground-truth 
and the BMVC 2009 system output. 

\subsection{Software}
The software implementation of GanitaMetrics 
is available for download from github at:
\url{https://github.com/GanitaOrg/ganitametrics.git} \cite{GM-url}.
Also, GanitaGraph needs to be installed. It is available at:
\url{https://github.com/GanitaOrg/ganitagraph.git}. 
The only major dependency for these packages is c++11
\footnote{GanitaMetrics and GanitaGraph do not depend on Boost.}. 

\section{Properties of the KL-inspired Tracking Metric}
In this section, we list properties of the tracking metric. 
We find it handles several use cases that are not handled collectively 
by previous metrics.  First, we give direct comparisons with MOTA 
using two major scenarios. This highlights the discontinuous nature 
of MOTA, based on the IoU setting of 0.5. Also, it shows that the score 
for MOTA can depend heavily on the raw number of frames in a video track, 
rather than on the relative length of the tracks. 

We also define 13 different simulations and give 
comparisons between our $D_{TD}$ metric and MOTA. 
We describe other properties of our metric, 
especially for certain edge cases. 
The final section highlights the main advantages 
of $D_{TD}$ over other video tracking metrics. 

\subsection{Comparison with MOTA}
\label{mota-compare}
We give two types of scenarios where 
the output from our KL-based metric varies 
significantly from the output of MOTA. 
First, note that MOTA is an accuracy measurement 
where higher scores are better, and $D_{TD}$ 
is an error measurement. 
Perfect scores for matching reference and system output are:
\begin{align*}
MOTA &= 1 \\ 
D_{TD} &= 0.
\end{align*}

\subsubsection{Split Tracks}
Suppose the reference consists of two disjoint 
tracks of $2n$ frames for each track. 
Suppose the system outputs four disjoint tracks 
of $n$ frames each, such that each reference 
is split into two system tracks with $n$ frames each. 
Then $D_{TD} = 1$.  However, for large $n$, 
$MOTA \approx 1$.  The only errors for MOTA 
originate from two ID switches, out of $4n$ frames total. 
This is a significant difference between these two metrics. 

Table \ref{table-split-all} 
shows the value of various 
metrics for the case of ten disjoint reference tracks, 
each with 100 frames, 
where half of the tracks split half way through. 
A python implementation (\cite{BES2006, cheind}) 
of the MOT metrics was used to 
produce the scores; other than $D_{TD}$ which was 
produced using our C++ implementation. 
\begin{table}[]
\centering
\small 
\begin{tabular}{|l|c|}
\cline{1-2}
Metric & Value \\
\cline{1-2}
MOTA & 0.995\\
MOTP & 0.000\\
IDR & 0.750\\
IDP & 0.750\\
IDF1 & 0.750\\
Recall & 1.000\\
Prec. & 1.000\\
MT & 10\\
ML & 0\\
FP & 0\\
FN & 0\\
IDs & 5\\
\cline{1-2}
$D_{TD}$ & 1.000\\
\cline{1-2}
\end{tabular}

\caption{Splitting Errors}
\label{table-split-all}
\end{table}

\subsubsection{Merged Tracks}
Suppose the reference has two tracks 
of equal size (same number of frames and 
same sized rectangles).  Suppose 
a system merges these two tracks 
such that each reference rectangle intersects 
exactly half of each system rectangle. 
Thus, the IoU for reference over system is exactly 
0.5 for each frame. 
Figure \ref{fig:merge-1} is a pictorial of two reference tracks, 
one red and the other blue, and a single system track in yellow. 
The horizontal black lines represent time. 

\begin{figure}
\centering
\begin{tikzpicture}[scale=0.6]
{[line width=2pt]
{[black]
\draw (3+.5,3.0) -- (3+9.5,3.0);
\draw (3+.5,3.8) -- (3+9.5,3.8);
}}

{[line width=1pt]
{[black]
    \filldraw (3,3) node [below] {} circle (1pt);
    \node [draw, thick, shape=rectangle, minimum width=0.6cm, minimum height=0.48cm, anchor=center, fill=red] at (3,3) {};
    \node [draw, thick, shape=rectangle, minimum width=0.6cm, minimum height=0.48cm, anchor=center, fill=red] at (5,3) {};
    \node [draw, thick, shape=rectangle, minimum width=0.6cm, minimum height=0.48cm, anchor=center, fill=red] at (7,3) {};
    \node [draw, thick, shape=rectangle, minimum width=0.6cm, minimum height=0.48cm, anchor=center, fill=red] at (9,3) {};
    \node [draw, thick, shape=rectangle, minimum width=0.6cm, minimum height=0.48cm, anchor=center, fill=red] at (11,3) {};
    \node [draw, thick, shape=rectangle, minimum width=0.6cm, minimum height=0.48cm, anchor=center, fill=red] at (13,3) {};

    \node [draw, thick, shape=rectangle, minimum width=0.6cm, minimum height=0.48cm, anchor=center, fill=blue] at (3,3.8) {};
    \node [draw, thick, shape=rectangle, minimum width=0.6cm, minimum height=0.48cm, anchor=center, fill=blue] at (5,3.8) {};
    \node [draw, thick, shape=rectangle, minimum width=0.6cm, minimum height=0.48cm, anchor=center, fill=blue] at (7,3.8) {};
    \node [draw, thick, shape=rectangle, minimum width=0.6cm, minimum height=0.48cm, anchor=center, fill=blue] at (9,3.8) {};
    \node [draw, thick, shape=rectangle, minimum width=0.6cm, minimum height=0.48cm, anchor=center, fill=blue] at (11,3.8) {};
    \node [draw, thick, shape=rectangle, minimum width=0.6cm, minimum height=0.48cm, anchor=center, fill=blue] at (13,3.8) {};
}}

{[line width=4pt]
{[yellow]
    \node [draw, line width=3pt, shape=rectangle, minimum width=0.6cm, minimum height=0.96cm, anchor=center] at (3,3.4) {};
    \node [draw, line width=3pt, shape=rectangle, minimum width=0.6cm, minimum height=0.96cm, anchor=center] at (5,3.4) {};
    \node [draw, line width=3pt, shape=rectangle, minimum width=0.6cm, minimum height=0.96cm, anchor=center] at (7,3.4) {};
    \node [draw, line width=3pt, shape=rectangle, minimum width=0.6cm, minimum height=0.96cm, anchor=center] at (9,3.4) {};
    \node [draw, line width=3pt, shape=rectangle, minimum width=0.6cm, minimum height=0.96cm, anchor=center] at (11,3.4) {};
    \node [draw, line width=3pt, shape=rectangle, minimum width=0.6cm, minimum height=0.96cm, anchor=center] at (13,3.4) {};
}
}
\end{tikzpicture}
\caption{Merged Tracks, $IoU=0.5$}
\label{fig:merge-1}
\end{figure}
The MOTA and $D_{TD}$ scores are:
\begin{align*}
MOTA &= 1 - \frac{ \sum_{t} \big( m_t + fp_t + \varPhi_t \big) }{ \sum_t g_t} \\
&= 1 - \frac{ \sum_{t} \big( 1 + 0 + 0 \big) }{ \sum_t 2 } = 0.5 
\end{align*}
\begin{align*}
D_{TD} &= 1\ \ \mbox{(inner div relative to system output)} 
\end{align*}

\begin{figure}
\centering
\begin{tikzpicture}[scale=0.6]
{[line width=2pt]
{[black]
\draw (3+.5,3.0) -- (3+9.5,3.0);
\draw (3+.5,3.8) -- (3+9.5,3.8);
}}

{[line width=1pt]
{[black]
    \filldraw (3,3) node [below] {} circle (1pt);
    \node [draw, thick, shape=rectangle, minimum width=0.6cm, minimum height=0.48cm, anchor=center, fill=red] at (3,3) {};
    \node [draw, thick, shape=rectangle, minimum width=0.6cm, minimum height=0.48cm, anchor=center, fill=red] at (5,3) {};
    \node [draw, thick, shape=rectangle, minimum width=0.6cm, minimum height=0.48cm, anchor=center, fill=red] at (7,3) {};
    \node [draw, thick, shape=rectangle, minimum width=0.6cm, minimum height=0.48cm, anchor=center, fill=red] at (9,3) {};
    \node [draw, thick, shape=rectangle, minimum width=0.6cm, minimum height=0.48cm, anchor=center, fill=red] at (11,3) {};
    \node [draw, thick, shape=rectangle, minimum width=0.6cm, minimum height=0.48cm, anchor=center, fill=red] at (13,3) {};

    \node [draw, thick, shape=rectangle, minimum width=0.6cm, minimum height=0.48cm, anchor=center, fill=blue] at (3,3.8) {};
    \node [draw, thick, shape=rectangle, minimum width=0.6cm, minimum height=0.48cm, anchor=center, fill=blue] at (5,3.8) {};
    \node [draw, thick, shape=rectangle, minimum width=0.6cm, minimum height=0.48cm, anchor=center, fill=blue] at (7,3.8) {};
    \node [draw, thick, shape=rectangle, minimum width=0.6cm, minimum height=0.48cm, anchor=center, fill=blue] at (9,3.8) {};
    \node [draw, thick, shape=rectangle, minimum width=0.6cm, minimum height=0.48cm, anchor=center, fill=blue] at (11,3.8) {};
    \node [draw, thick, shape=rectangle, minimum width=0.6cm, minimum height=0.48cm, anchor=center, fill=blue] at (13,3.8) {};
}}

{[line width=4pt]
{[yellow]
    \node [draw, line width=3pt, shape=rectangle, minimum width=0.6cm, minimum height=1.2cm, anchor=center] at (3,3.4) {};
    \node [draw, line width=3pt, shape=rectangle, minimum width=0.6cm, minimum height=1.2cm, anchor=center] at (5,3.4) {};
    \node [draw, line width=3pt, shape=rectangle, minimum width=0.6cm, minimum height=1.2cm, anchor=center] at (7,3.4) {};
    \node [draw, line width=3pt, shape=rectangle, minimum width=0.6cm, minimum height=1.2cm, anchor=center] at (9,3.4) {};
    \node [draw, line width=3pt, shape=rectangle, minimum width=0.6cm, minimum height=1.2cm, anchor=center] at (11,3.4) {};
    \node [draw, line width=3pt, shape=rectangle, minimum width=0.6cm, minimum height=1.2cm, anchor=center] at (13,3.4) {};
}
}
\end{tikzpicture}
\caption{Merged Tracks, $IoU < 0.5$}
\label{fig:merge-2}
\end{figure}
Note, if the system outputs a single merged track such that each reference rectangle 
is slightly less than half of each system rectangle, then the MOTA score becomes negative. 
In particular, 
\begin{align*}
MOTA &= 1 - \frac{ \sum_{t} \big( m_t + fp_t + \varPhi_t \big) }{ \sum_t g_t} \\
&= 1 - \frac{ \sum_{t} \big( 2 + 1 + 0 \big) }{ \sum_t 2 } = - 0.5 
\end{align*}
However, since $D_{TD}$ is continuous, the score does not change much 
from the previous case and is still close to 1. 
\begin{align*}
D_{TD} &= 1.003690\ \ \mbox{(inner div + false alarm error)} 
\end{align*}
Figure \ref{fig:merge-2} shows a pictorial of this scenario. 

\subsection{Tracking scenarios}
We present error results under 13 different scenarios.  Three different 
scenarios for truthed tracks are used, and then multiple scenarios 
for simulated system output are compared.
\begin{enumerate}
\itemsep-.10em 
\item[T1.] Truth track set 1 contains two tracks with 5 frames each. 
The tracks intersect exactly on the 3rd frame. 
\item[T2.] Truth track set 2 contains two disjoint tracks with 5 frames each. 
\item[T3.] Truth track set 3 contains ten disjoint tracks with 10 frames each.
\end{enumerate}
For each simulated system output (S1 - S13), 
we computed the error, as compared to the given ground-truth scenario 
(T1, T2 or T3). 
In Table \ref{table3}, the simulated tracks are listed 
in order from the lowest error to the highest error. 
For further details, see the GanitaMetrics software 
package \cite{GM-url}, 
where simulated tracks are provided for all of these scenarios. 
This is the list of simulated scenarios (S1 - S13):
\begin{enumerate} 
\label{list1}
\itemsep-.10em 
\item[S1.] Control case where system output equals truth
\item[S2.] One track is correct; other flips half way
\item[S3.] Tracks switch after intersecting in the middle
\item[S4.] Instead of 2 tracks, system splits into 4 tracks
\item[S5.] System gets 40\% of 1 track and 60\% of the other
\item[S6.] One system track is missing 40\% of track
\item[S7.] Missing 1 of the tracks
\item[S8.] One track is generated twice
\item[S9.] System outputs half of each truthed box for each track
\item[S10.] System outputs first half of track temporally
\item[S11.] System misses half of the tracks (no intersection)
\item[S12.] System gets 7 of 10 tracks; misses other 3
\item[S13.] System gets 90\% of each track
\end{enumerate}
\begin{table}
\centering
\begin{tabular}{|c|c|c|c|}
\cline{1-4}
Truth set & Simulations & KL-track error & MOTA\\
\cline{1-4}
T1 & S1 & 0.000000 & 1.000000\\
T3 & S13 & 0.262899 & 0.864000\\
T1 & S6 & 0.392614 & 0.800000\\
T1 & S3 & 0.839946 & 0.800000\\
T1 & S4 & 0.970951 & 0.800000\\
T2 & S8 & 1.000000 & 0.500000\\
T1 & S2 & 1.024807 & 0.600000\\
T3 & S12 & 1.188722 & 0.700000\\
T1 & S5 & 1.204928 & 0.600000\\
T3 & S9 & 1.304112 & 1.000000\\
T3 & S10 & 1.304112 & 0.732000\\
T1 & S7 & 1.414153 & 0.500000\\
T3 & S11 & 2.339462 & 0.500000\\
\cline{1-4}
\end{tabular}
\caption{Error Summary for Simulated Scenarios}
\label{table3}
\end{table}

\subsection{Advantages, general properties and edge cases}
The list shown in Table \ref{adv} gives general properties or 
prospective advantages of our KL-based tracking metric. 
\begin{table}
\small 
\begin{enumerate}
\itemsep-.10em 
\item {\it No need to introduce thresholds (i.e., IoU);}
\item {\it Error measurements are continuous with respect to system tracks;}
\item {\it No need to enforce a one-to-one track alignment;}
\item {\it Metric is based on the distance between spatio-temporal volumes;}
\item {\it Split tracks produce lower error rates than false alarms;}
\item {\it Error levels do not necessarily tend to infinity with the number of reference tracks.}
\end{enumerate}
\caption{Properties of KL-inspired Metric}\label{adv}
\end{table}
Typically, when thresholds are introduced into tracking metrics, this causes 
discontinuities in the scoring.  Many times this can impact the rankings 
of the systems that are being measured.  Also, thresholds when applied 
to IoU
tend to incentivize focus at a specific parameter level, while leading 
to weaker performance toward the overall 
goal of tracking.  Since our metric does not introduce arbitrary thresholds, 
the metric is continuous with respect to system tracks (as well as ground-truth tracks). 

A key property of KL-divergence is the use of the function 
$f(x) = - x \log{(x)}$ for computing entropy or counting 
discrepancies between discrete distributions.  
Since 
\[
\lim_{x\to 0^+} x \log{x} = 0 , 
\]
then $f(x)$ may be extended to a continuous function 
on $[0,1]$ such that $f(0)=0$, and also $f(1) = 0$. 
Thus, for split tracks, the error goes to zero 
as the overlap shrinks to zero or as the overlap 
converges to 1.  A max occurs at 
$x = {1} / {(e \ln{2})}$. 
The reduction in error for larger overlaps occurs 
without the need to program specific thresholds for temporal overlap. 
See Table \ref{table3} for a comparison of errors in various scenarios.

Other characteristics of this metric is that the overall score does not 
necessarily depend strongly on the number of reference tracks. 
In particular, certain scenarios produce a similar overall error, regardless 
of the number of tracks.  Suppose that there are $n$ disjoint reference tracks 
of equal length. 
Suppose a system generates $n$ disjoint system tracks, and for each reference track, 
there is a single system track that overlaps it by half. 
In this case, our metrics generate error in two places: 
inner divergence relative to reference, and outer divergence relative to reference. 
For the inner divergence error, we get $0.5 \times \log{(2)}$ error for each 
reference track.  Then a uniform mean is computed to give a final inner divergence 
error of $0.5 \times \log{(2)}$. 
The outer divergence relative to a reference track is 
\[
\log{ \Big( \frac{2 + n}{1 + \frac{1}{2}(1+n)} \Big) } = 
\log{ \Big( 2 \big( \frac{2 + n}{3 + n}\big) \Big) } < \log{(2)} . 
\]
To get the overall outer divergence relative to the reference track set, 
these errors are averaged uniformly over all reference tracks. 
In this case, the total error will approach $1.5 \log{(2)}$, as $n\to \infty$. 

%
Although, no thresholds are used to generate errors, design choices were made 
to formulate the KL-divergence in this setting.  
We found that errors were not generated in an obviously counter-intuitive manner. 
For example, systems that split a reference track produced less error than another 
system that moved the split piece to be a false alarm. 
A consumer of this metric might be interested in penalizing 
false alarms more heavily than missed detections. 
This could be accomplished readily by introducing weights into the 
general formula in section \ref{gen-for-subsection}. 
It would still produce a continuous metric and satisfy 
most of our other goals.

\subsection{Bounded error scenario}
\label{bes-subsection}
Our KL-based metric penalizes systems that completely miss tracks at a higher 
cost than systems that overlap tracks non-negligibly. 
Suppose there are $n$ disjoint ground-truth tracks, 
and a system outputs $n$ system tracks such that each system track 
intersects one and only one ground-truth track at a random 
proportion $x \in (0,1)$.  In this case, the missed detection error 
is bounded almost everywhere, and the error converges to ${1} / {\ln{(2)}}$ 
as $n\to \infty$. 
This is due to the law of large numbers and the fact that 
\begin{eqnarray*}
\int_0^1 \log{ \Big( \frac{2+n}{1+(1+n)x} \Big) } dx &=& 
\frac{1}{\ln{(2)}} \Big( 1 - \frac{\ln{(2+n)}}{1+n} \Big) \\ 
&\to& \frac{1}{\ln{(2)}} 
= 1.442695\ldots . 
\end{eqnarray*}
Note, this integral can be computed using integration by parts. 

\subsection{Activity detection}
An area of research that is seeing increased focus 
is detection of events or activities in video. 
This builds on frame-based object detection and recognition. 
The metric described in this paper could be used to measure 
the accuracy of activity detection. It provides a holistic 
measure for comparing two sets of spatio-temporal volumes. 

There have been several challenges in the area of activity recognition including 
\cite{Patino_2017_CVPR_Workshops, Charades2016, HEG2015, RFOGM2008}. 
The focus is on classification or recognition in short untrimmed video segments.  
Recently, the ActEV (Activities in Extended Video) challenge 
\cite{ActEV2018} has looked at fully unconstrained activity detection 
in surveillance video. 
Still, the metric described in this paper is fundamentally different 
from existing activity detection metrics; existing metrics require 
prescribed thresholds, do not provide a continuous measure 
on spatio-temporal volumes, or rely on one-to-one correspondences between 
system output and ground-truth detections. 



\section{Conclusion}
We have defined a new 
spatio-temporal tracking metric 
which is a continuous, parameter-less, full-scope metric. 
Also, we give direct comparisons with established metrics, 
and demonstrate typical scenarios where our KL-based metric 
gives a more intuitive and consistent score than the most 
common tracking metric in use today.
A software implementation of this metric is available 
at \cite{GM-url}. 


\section*{Acknowledgment}
The author would like to thank Dr. Reuven Meth. 
Also, the author wishes to thank Dr. Yonatan Tariku 
for providing challenging tracking scenarios 
and data (i.e., section \ref{exp-with-means}).  
This was used to test and improve the overall tracking metric.

{\small
\bibliographystyle{amsalpha}
\bibliography{mybib}
}

\end{document}